%% file: acl_latex.tex
\pdfoutput=1

\documentclass[11pt]{article}

\usepackage[final]{acl}

\usepackage{times}
\usepackage{latexsym}
\usepackage{multirow}
\usepackage{booktabs}
\usepackage{makecell}
\usepackage[normalem]{ulem}
\usepackage{amsmath}
\usepackage{CJKutf8}

\usepackage[T1]{fontenc}

\usepackage[utf8]{inputenc}

\usepackage{microtype}

\usepackage{inconsolata}

\usepackage{graphicx}

\title{Do Large Language Models Judge Error Severity Like Humans?}

\title{Do Large Language Models Judge Error Severity Like Humans?}
\author{
Diege Sun\textsuperscript{1}, Guanyi Chen\textsuperscript{2,3,4\thanks{Corresponding Authors}}, Zhao Fan\textsuperscript{1$^*$}, Xiaorong Cheng\textsuperscript{1},  Tingting He\textsuperscript{2,3,4}   \\
\textsuperscript{1}School of Psychology,\\
\textsuperscript{2}Hubei Provincial Key Laboratory of Artificial Intelligence
and Smart Learning,\\
\textsuperscript{3}National Language Resources Monitor and Research Center
for Network Media,\\
\textsuperscript{4}School of Computer Science, Central China Normal
University \\
\texttt{\{g.chen, zfan\}@ccnu.edu.cn}
}

\begin{document}
\maketitle
\begin{abstract}
Large Language Models (LLMs) are increasingly used as automated evaluators in natural language generation, yet it remains unclear whether they can accurately replicate human judgments of error severity. In this study, we systematically compare human and LLM assessments of image descriptions containing controlled semantic errors. We extend the experimental framework of~\citet{van-miltenburg-etal-2020-gradations} to both unimodal (text-only) and multimodal (text + image) settings, evaluating four error types: age, gender, clothing type, and clothing colour. Our findings reveal that humans assign varying levels of severity to different error types, with visual context significantly amplifying perceived severity for colour and type errors. Notably, most LLMs assign low scores to gender errors but disproportionately high scores to colour errors, unlike humans, who judge both as highly severe but for different reasons. This suggests that these models may have internalised social norms influencing gender judgments but lack the perceptual grounding to emulate human sensitivity to colour, which is shaped by distinct neural mechanisms. Only one of the evaluated LLMs, Doubao, replicates the human-like ranking of error severity, but it fails to distinguish between error types as clearly as humans. Surprisingly, DeepSeek-V3, a unimodal LLM, achieves the highest alignment with human judgments across both unimodal and multimodal conditions, outperforming even state-of-the-art multimodal models.
\end{abstract}

\begin{CJK*}{UTF8}{gbsn}
\input{section/intro}
\input{section/hypo}

\input{section/experiment}
\input{section/result}
\input{section/discussion}

\section{Conclusion}

This study investigates whether large language models (LLMs) judge the severity of errors in image descriptions in a manner consistent with human evaluators. Building upon the framework of \citet{van-miltenburg-etal-2020-gradations}, we designed a controlled experiment comparing human and LLM assessments across unimodal and multimodal conditions. Our findings show that while humans demonstrate clear and nuanced severity distinctions across different error types, especially influenced by visual context in the case of colour and type errors, LLMs struggle to fully replicate these patterns.

Interestingly, several LLMs, including GPT-4o and the DeepSeek models, assign lower scores to gender errors but surprisingly high scores to colour errors, diverging from human evaluators. This discrepancy may stem from the fact that humans perceive colour errors as severe due to specialised neural mechanisms for processing colour, whereas gender errors are judged more harshly due to socially embedded norms. The LLMs appear to have internalised aspects of social norms but lack the perceptual grounding needed to emulate the human sensitivity to colour-related inaccuracies.

Among the evaluated models, Doubao was the only LLM to reproduce the human-like severity ranking, though it did not consistently distinguish severity across error types. In contrast, DeepSeek-V3 emerged as the best overall LLM for serving as an NLG evaluator. It achieved the highest correlation with human judgments across both unimodal and multimodal inputs, and it did so with greater cost efficiency than the multimodal models. Despite overestimating the acceptability of colour errors, DeepSeek-V3 demonstrated the most consistent alignment with human assessments.

\section*{Limitations}

One limitation of this study is that it includes two closed-source LLMs, Doubao and GPT-4o, whose underlying model architectures and evaluation pipelines are not publicly documented. This raises the possibility that different internal models or processing strategies may be used for unimodal and multimodal inputs, potentially affecting the validity of some interaction effect tests in this study. Nevertheless, this limitation does not undermine the methodology we propose. Our experimental framework remains broadly applicable and can be readily used to evaluate other multimodal LLMs in controlled settings, particularly when the models can be deployed locally with full transparency.

\bibliography{anthology,custom}


\appendix

\input{section/appendix}


\end{CJK*}

\end{document}

%% file: section/intro.tex
\section{Introduction}

Large Language Models (LLMs) are increasingly being explored not only as generators of text~\citep{xuanfan-piji-2023-systematic}, but also as evaluators, offering a scalable alternative to human judgment in Natural Language Generation (NLG) evaluation~\citep{gu2024survey,gao2025llm}. Their use as ``LLM-as-a-judge'' has shown promise in areas such as summarisation and dialogue~\citep{liu-etal-2023-g,liusie-etal-2024-llm}, where traditional metrics, such as BLEU~\citep{papineni-etal-2002-bleu} and ROUGE~\citep{lin2004rouge}, often fall short of capturing semantic nuances and user preferences.

\begin{figure}
    \centering
    \includegraphics[width=0.4\linewidth]{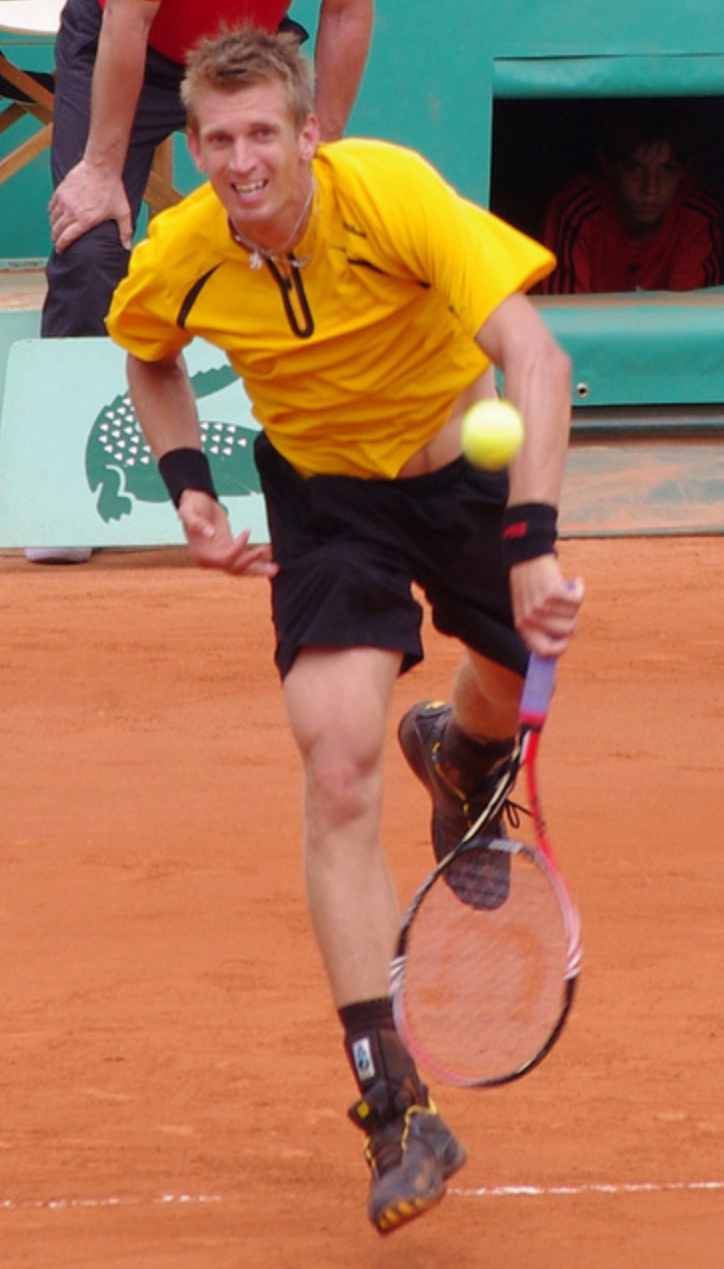}
    \caption{Image 34419 from MSCOCO: \emph{A man wearing a yellow shirt on a tennis court plays tennis.}}
    \label{fig:example}
\end{figure}

Nonetheless, key questions remain about whether LLMs can reliably mirror human judgment~\cite{bavaresco2024llms,gao2025llm}. One underexplored challenge is the assessment of error severity in generated text. \citet{van-miltenburg-etal-2020-gradations} observed that, in image captioning, human evaluators do not treat all errors equally: for example, given Figure~\ref{fig:example}, the caption ``\emph{a man wearing a \textbf{red} shirt}'' is judged more severely incorrect than ``\emph{a man wearing a yellow \textbf{coat}}''. Whether (multi-modal) LLMs’ assessments of error severity align with such human intuitions remains an open question.

\citet{van-miltenburg-etal-2020-gradations} focused on multimodal NLG evaluation, where human judges evaluate outputs with access to both text and visual input. They showed that humans assign different levels of severity to different types of errors, suggesting nuanced perceptions of correctness. However, many real-world NLG tasks, such as summarisation, dialogue, and data-to-text generation, are unimodal (text-only), and most current LLMs are also unimodal. This raises a critical question: Are the severity distinctions observed in multimodal evaluations dependent on the presence of visual context, or do they reflect more general patterns in human judgment? 

In other words, it remains unclear how severity judgments of human beings are formed in unimodal settings, and how the presence of visual context may influence those judgments. With these two questions about human severity judgement in mind, we are further interested in whether unimodal LLMs' severity judgement aligns with that of humans and whether visual context has a similar impact on multimodal LLMs.

We therefore come up with two core research questions:
\begin{itemize}
    \item \textbf{Do Large Language Models judge error severity like humans?} We examine the extent to which LLM-based evaluations agree with human assessments across a range of error types in NLG.
    \item \textbf{How does the presence of images influence human judgment of errors in NLG?} We explore whether access to visual context alters human perceptions of error severity and whether this impact is mirrored in LLM-based assessments.
\end{itemize}

To address these two questions, we begin by updating the experimental setup of \citet{van-miltenburg-etal-2020-gradations}, allowing human judges to evaluate outputs in Chinese under two conditions: with access to both visual and textual inputs (multimodal), or with access to only textual inputs (unimodal). This design enables us to examine how severity judgments are formed in unimodal settings and to assess the influence of visual context. Next, we investigate the extent to which LLMs’ assessments of error severity align with those of humans by replicating the experiment using a range of unimodal and multimodal LLMs. Finally, we conduct a systematic comparison of human and LLM behaviour, evaluating which models best approximate human judgments of error severity across unimodal and multimodal conditions.

%% file: section/hypo.tex
\section{Hypotheses}

Following \citet{van2017room} and \citet{van-miltenburg-etal-2020-gradations}, we are interested in four types of errors in Chinese Image Descriptions: \textsc{age}, \textsc{gender}, \textsc{clothing-colour} (henceforth, \textsc{colour}), and \textsc{clothing-type} (henceforth, \textsc{type}) (See Appendix~\ref{sec:appendix_example} for examples of each error). The current section discusses the hypotheses of our experiment. 

\citet{van-miltenburg-etal-2020-gradations} asked participants to rate the quality of image descriptions, each containing a single type of error, with reference to the corresponding image. Lower ratings indicated more severe errors. Their findings revealed the following ranking of error types in terms of severity:
\begin{equation}\label{eq:order_1}
    \text{\textsc{colour}} \prec \text{\textsc{gender}} \prec \text{\textsc{type}} \prec \text{\textsc{age}}
\end{equation}
where $A \prec B$ means error type $A$ receives lower scores (i.e., was judged more severe) than $B$. Notably, the differences in severity between \textsc{colour} and \textsc{age}, \textsc{colour} and \textsc{type}, as well as \textsc{gender} and \textsc{age}, were found to be statistically significant. 

In this study, we allow participants to access either both visual and textual inputs or only textual inputs to explore the impact of visual context. Previous studies in cognitive science have shown that visual context influences the processing of colour perception~\citep{adelson1993perceptual,chen-etal-2019-generating}, age perception~\citep{pilz2022contextual}, object type recognition~\citep{fenske2006top}, and gender representation~\citep{guilbeault2024online} in the human brain. Based on these findings, we expect that the presence of visual context will significantly affect the ratings given by participants. However, we do not expect it to alter the severity ranking of errors observed in \citet{van-miltenburg-etal-2020-gradations}. For the LLM replications, considering the potential of LLMs to approximate human judgments, we hypothesise that these models will replicate human behaviour in our updated experiment. Formally, we formulate the following hypotheses:
\begin{description}
    \item[$\mathcal{H}_1$] Humans will reproduce the same severity ranking of error types as reported by \citet{van-miltenburg-etal-2020-gradations}, under both multimodal and unimodal conditions.
    \item[$\mathcal{H}_2$] The presence of visual context will have a significant effect on human judgments.
    \item[$\mathcal{H}_3$] Both multimodal and unimodal LLMs will replicate the same severity ranking of error types as observed in human judgments.
    \item[$\mathcal{H}_3$] Visual context will influence multimodal LLMs in a manner similar to its effect on human judgments.
\end{description}

%% file: section/experiment.tex
\section{Experiments}

We elaborate on experimental settings and stimuli used for our human and LLM experiments.

\subsection{Human Experiment}

\paragraph{Material.} We selected a total of 9 images, 5 of which are sourced from MS COCO~\citep{lin2014microsoft}. To prevent data contamination~\citep{balloccu-etal-2024-leak} in the subsequent LLM experiment, the remaining 4 images were sourced from the internet. These images were chosen following the same criteria outlined by \citet{van-miltenburg-etal-2020-gradations} to minimise ambiguity in error types (for further details, see Appendix~\ref{sec:appendix_criteria}). This ensures that, for example, when a description refers to a person in the image as ``\emph{young}'', the individual in the image is unmistakably young.

Following the approach of \citet{van-miltenburg-etal-2020-gradations}, we tasked two native speakers of Mandarin Chinese with writing a reference description for each image, as well as creating minimal pairs between erroneous descriptions and the reference descriptions. Specifically, the annotators were instructed to generate erroneous descriptions based on four error types: \textsc{age}, \textsc{gender}, \textsc{colour}, and \textsc{type}, by altering a single character in the Chinese text. To minimise the influence of vagueness, the annotators were asked to ensure that the difference between the reference and erroneous descriptions was substantial. For instance, they were instructed to change ``red'' to ``blue'' rather than to a more subtle colour like ``orange''. The two annotators discussed finalising the descriptions. As a result, each image is paired with one reference description and four erroneous descriptions. An example can be found in Appendix~\ref{sec:appendix_example}.

\begin{figure}[t]
    \centering
    \includegraphics[scale=0.5]{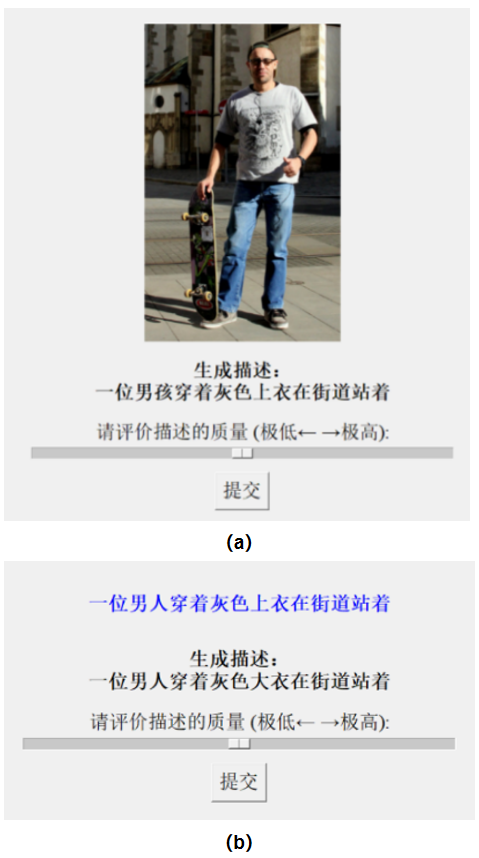}
    \caption{Screenshots of a trial in (a) multimodal condition, and (b) unimodal condition.}
    \label{fig:screenshot}
\end{figure}

\begin{figure*}[t]
    \centering
    \includegraphics[scale=0.35]{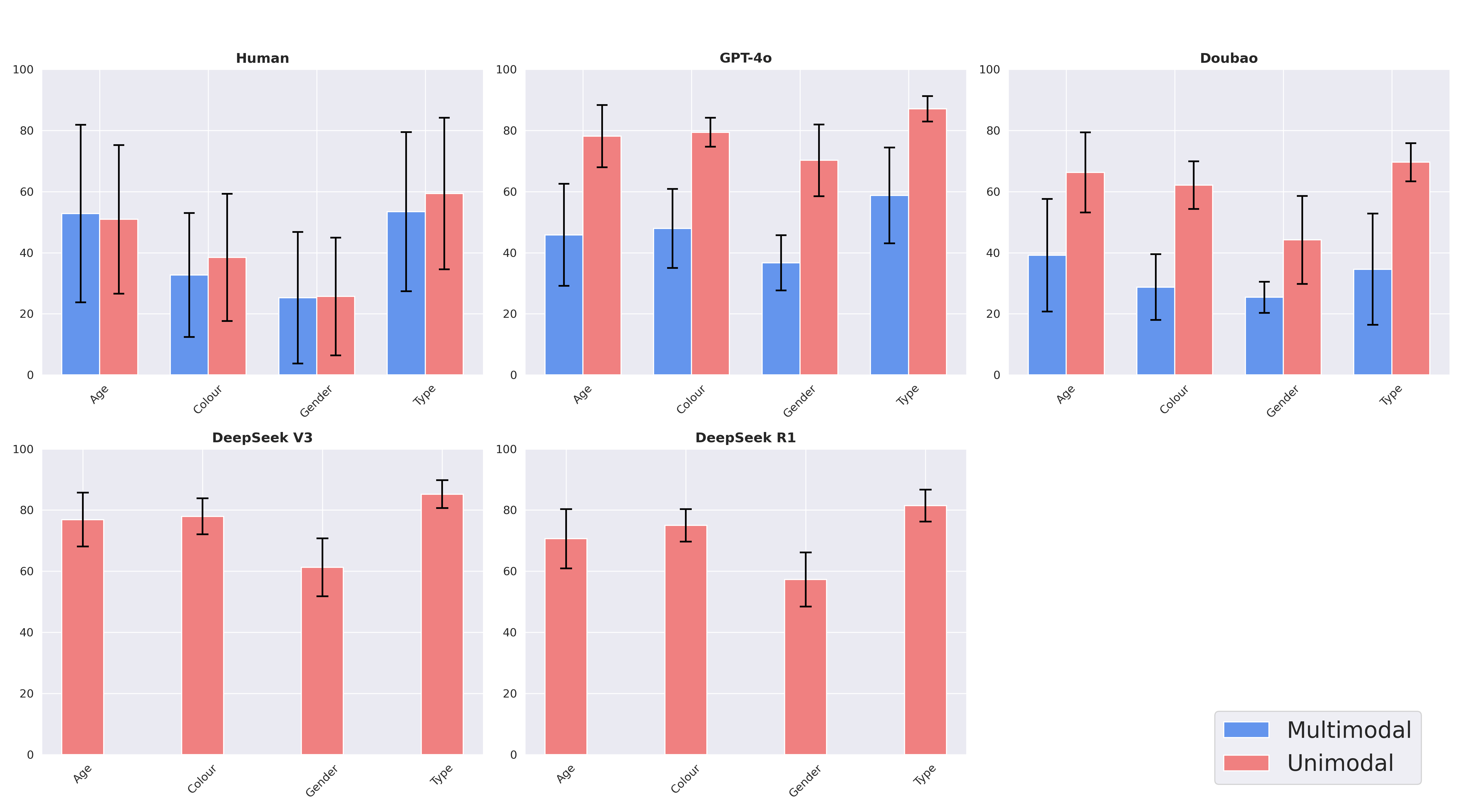}
    \caption{The results of both our human experiment and LLM experiment. Each error bar indicates the standard deviation of scores.}
    \label{fig:score}
\end{figure*}

\paragraph{Design.} The stimuli in our experiment consist of two conditions: multimodal and unimodal. In the multimodal condition, participants were shown an image alongside a corresponding erroneous description. In the unimodal condition, participants were shown a reference description along with a corresponding erroneous description. This results in a total of $9\times4\times2=72$ stimuli. It is important to note that our experimental design in the multimodal condition differs from that of \citet{van-miltenburg-etal-2020-gradations}. Specifically, we did not present the reference description in the multimodal condition to avoid potential bias in participants' decisions caused by prior exposure to the reference text.

Stimuli of one image from MS COCO were used as the practice for participants. We thus have 64 stimuli in the main experiment. In each trial, participants rated the quality of the erroneous description using a slider ranging from 0 to 100 (i.e., Magnitude Estimation; ~\citet{bard1996magnitude}). Screenshots of both conditions are shown in Figure~\ref{fig:screenshot}. 

\paragraph{Procedure.} Participants were invited to the lab to take part in the experiment. The session began with an instruction informing them that their task was to rate the quality of a set of descriptions generated by an artificial intelligence system, based on either reference images or reference texts. Participants were told that the ratings would be done using a slider ranging from 0 to 100, where a higher score indicated better quality. They were not informed that the descriptions they would rate contained errors.

The experiment began with a demographic survey, where participants provided information about their age, gender, educational background, and whether they had colour blindness. Following this, participants moved on to the practice phase, where they rated 8 practice stimuli corresponding to the same image. These 8 practice stimuli were shuffled randomly and were identical for all participants.

After completing the practice phase, participants were asked to press the ``START" button when they fully understood the experimental procedure and were ready to proceed to the main experiment. The stimuli in the main experiment were divided into two groups. Each group contained 32 stimuli: 16 stimuli of 4 images in the multimodal condition and 16 stimuli of the other 4 images in the unimodal condition, with stimuli shuffled randomly within each group. The main experiment consisted of three phases. In the first phase, participants rated the stimuli from the first group without interruption. The second phase was a rest period, during which participants were asked to press the ``START" button when they were ready to continue. In the third phase, participants rated the stimuli from the second group. During the main experiment, we recorded each participant's ratings for each trial, as well as their reaction times.

\paragraph{Participants.} We used G*Power to estimate the required sample size for our experiment. Since we planned to use a Linear Mixed-Effects Model (LMM) for statistical analysis (see Section~\ref{sec:result} for further details), we set the effect size measure (Cohen's $f$) to 0.23, the significance level ($\alpha$) to 0.05, and the statistical power ($1-\beta$) to 0.8. Based on the number of stimuli, the required sample size was estimated to be 18 participants. To mitigate potential ceiling or floor effects, we ultimately collected data from 25 participants. Of these, 14 were male and 11 were female. The participants had a mean age of 22.92 years (SD = 3.52). All participants were university students with higher education backgrounds. Prior to the experiment, all participants confirmed that they had no history of colour blindness or colour vision deficiencies.

\subsection{LLM Experiment}

\input{table/effect}
\input{table/fee}

We used several unimodal and multimodal LLMs to replicate the above experiment to explore their error severity judgements.

\paragraph{Material and Design.} The same 64 stimuli used in the human experiment were also employed in the LLM experiment. We instructed the LLMs to assign each stimulus a quality score ranging from 0 to 100. For unimodal stimuli, we used the prompt in Table~\ref{tab:prompt_uni} (in Appendix~\ref{sec:appendix_prompt}) to guide both unimodal and multimodal LLMs in rating these stimuli. For multimodal stimuli, we used the prompt in Table~\ref{tab:prompt_mul} to instruct multimodal LLMs to rate the stimuli. In both prompts, we also requested that the LLMs provide an explanation for the scores they assigned.

\paragraph{Procedure and Models.} We selected two powerful multimodal LLMs, GPT-4o~\citep{hurst2024gpt} and Doubao\footnote{\url{https://seed.bytedance.com/en/special/doubao_1_5_pro}}, and two Unimodal LLMs, DeepSeek-V3~\citep{liu2024deepseek}\footnote{We used the DeepSeek-V3-0324.} and DeepSeek-R1~\citep{guo2025deepseek}. We included DeepSeek-R1 to investigate whether reasoning LLMs are better at distinguishing the severity of different types of errors. For each trial, we run each LLM 3 times to mitigate the influence of randomness.

%% file: table/effect.tex
\begin{table}[t]
\resizebox{\columnwidth}{!}{%
\centering
\begin{tabular}{llllll}
\toprule
& Human & GPT-4o & Doubao \\
\midrule
Error Type   & $<.001^{***}$ & $<.001^{***}$ & $<.001^{***}$ \\
Visual Context & $.002^{**}$ & $<.001^{***}$ & $<.001^{***}$ \\
Interaction Effect & $.014^{*}$ & $.677$ & $.004^{**}$ \\ 
\bottomrule
\end{tabular}}
\caption{The LMM-based statistical test results for the main effects of error type and visual context, as well as their interaction effect. Significance Marking: $^{***} p<.001$; $^{**} p<.01$; $^{*} p<.05$.}
\label{tab:effect}
\end{table}


%% file: table/fee.tex
\begin{table*}[t]
\small
\centering
\begin{tabular}{lrrlrrlrrl}
\toprule
\multirow{2}{*}{} & \multicolumn{3}{c}{Human}    & \multicolumn{3}{c}{GPT-4o}   & \multicolumn{3}{c}{Doubao}\\
\cmidrule(lr){2-4} \cmidrule(lr){5-7}\cmidrule(lr){8-10}
& Est. & t & p-value & Est. & t & p-value & Est. & t & p-value \\
\midrule
(Intercept)  & 52.795 & 19.777 & $<.001^{***}$ & 45.833 & 17.831 & $<.001^{***}$ & 39.167 & 13.937 & $<.001^{***}$ \\ \midrule
Colour Error & -20.085 & -9.679 & $<.001^{***}$ & 2.083 & 0.658 & .512 & -10.417 & -2.957 & $.003^{**}$ \\
Type Error & 0.625 & 0.301 & $.763$ & 12.917 & 4.077 & $<.001^{***}$ & -4.583 & -1.301 & .195 \\
Gender Error & -27.570 & -13.287 & $<.001^{***}$ & -9.167 & -2.893 & $.004^{**}$ & -13.750 & -3.903 & $<.001^{***}$ \\ \midrule
Unimodal & -1.895 & -0.913 & $.361$ & 32.292 & 10.191 & $<.001^{***}$ & 27.083 & 7.688 & $<.001^{***}$ \\ \midrule
Colour $\times$ Unimodal & 7.635 & 2.602 & $.009^{**}$ & -0.833 & -0.186 & .853 & 6.250 & 1.255 & .211 \\
Type $\times$ Unimodal & 7.800 & 2.658 & $.008^{**}$ & -3.958 & -0.883 & .378 & 7.917 & .113 & .113 \\
Gender $\times$ Unimodal & 2.325 & 0.792 & $.428$ & 1.250 & 0.279 & .781 & -8.333 & -1.673 & .096 \\
\bottomrule
\end{tabular}
\caption{The estimated fixed effects in LMM for Human, GPT-4o and Doubao. It uses the age error in the multimodal condition as the reference group (i.e., intercept). `Est.' is the estimated fixed effect coefficient, and `t' is the t-value.}
\label{tab:fee}
\end{table*}

%% file: section/result.tex
\section{Results} \label{sec:result}


\input{table/post_hoc}

Figure~\ref{fig:score} reports the scores given by humans and 5 LLMs. 

\subsection{Human Performance}

Hypothesis $\mathcal{H}_1$ predicted that the severity ranking of error types would align with the ranking reported by \citet{van-miltenburg-etal-2020-gradations} in both multimodal and unimodal conditions. The results of our human experiment show that, while the ranking is consistent across the two conditions, it differs from the one reported by \citet{van-miltenburg-etal-2020-gradations}:
\begin{equation} \label{eq:order_2}
    \text{\textsc{gender}} \prec \text{\textsc{colour}} \prec \text{\textsc{age}} \prec \text{\textsc{type}}
\end{equation}
Thus, we must reject hypothesis $\mathcal{H}_1$. A potential explanation for this discrepancy is that our experimental settings differ slightly from those of \citet{van-miltenburg-etal-2020-gradations} in two key ways: (1) we conducted an in-lab experiment, while they used a crowdsourcing approach, and (2) in the multimodal condition, we showed participants only the image and the erroneous description to minimize bias, whereas \citet{van-miltenburg-etal-2020-gradations} also presented the reference description.

\paragraph{Effects of Error Type and Visual Context.}

The results suggest that participants assigned varying quality scores across different error types. Additionally, the presence of visual context had a slight influence on the scores participants assigned, though it did not alter the severity ranking. To ascertain the influence of these two factors, we conduct a statistical analysis using a Linear Mixed-Effects Model (LMM). \footnote{We chose an LMM because, although the default position of the slider in the experiment is in the middle, the resulting scores are not normally distributed. This was confirmed using a Shapiro-Wilk test, which indicated that the distribution of scores significantly deviates from a normal distribution ($W=0.959, p<.001$).} The outcomes of the analysis using LMM are reported in Table~\ref{tab:effect}, which shows that both the error type and the presence of visual context have significant main effects on human judgements. Additionally, the interaction effect between these two factors is significant, indicating that their impacts are dependent on each other.

To further investigate how the presence of visual context influences human judgments, we report the estimated fixed effects in Table~\ref{tab:fee}. Interestingly, the fixed effect estimates indicate that, overall, the presence of visual context does not significantly alter the quality scores assigned by human participants ($p=.361$; see the `Unimodal' line in Table~\ref{tab:fee}). However, it does have a significant main effect on human judgments due to interaction effects on colour errors ($p=.009$) and type errors ($p=.008$; see the last three lines in Table~\ref{tab:fee}). In other words, the answer to the hypothesis $\mathcal{H}_2$ is that the presence of visual context significantly influences human judgments of error severity, but only for colour and type errors, making participants perceive these two errors as more severe.

\paragraph{Pairwise Comparisons.} We extracted the estimated marginal means from the Linear Mixed-Effects Model (LMM) to perform pairwise comparisons between different error types. The estimated differences and their corresponding significance levels are reported in Table~\ref{tab:post_hoc}. These pairwise comparisons were conducted both for the main effects (i.e., comparing the overall quality scores between pairs of error types) and for the interaction effects (i.e., comparing the quality scores separately in the unimodal and multimodal conditions). To determine the significance of these pairwise comparisons, we applied Tukey's Honestly Significant Difference (HSD) test~\citep{tukey1949comparing} for the main effects and the Holm–Bonferroni method~\citep{holm1979simple} for the interaction effects.

The results indicate that almost all differences between scores for different error types are significant. This is a bit different from the findings of \citet{van-miltenburg-etal-2020-gradations}, where half of the differences were found to be statistically indistinguishable. One possible explanation for this discrepancy is that the in-lab setting of our experiment likely enhanced participants' focus, enabling them to identify errors more easily. The only exception to this pattern is the difference between age and type errors in the multimodal condition, which was not significant. This may be attributed to the presence of visual context, which, as mentioned earlier, led participants to perceive type errors as more severe, thereby reducing the difference between type and age errors.

\paragraph{Reaction Time}

\begin{figure}
    \centering
    \includegraphics[scale=0.35]{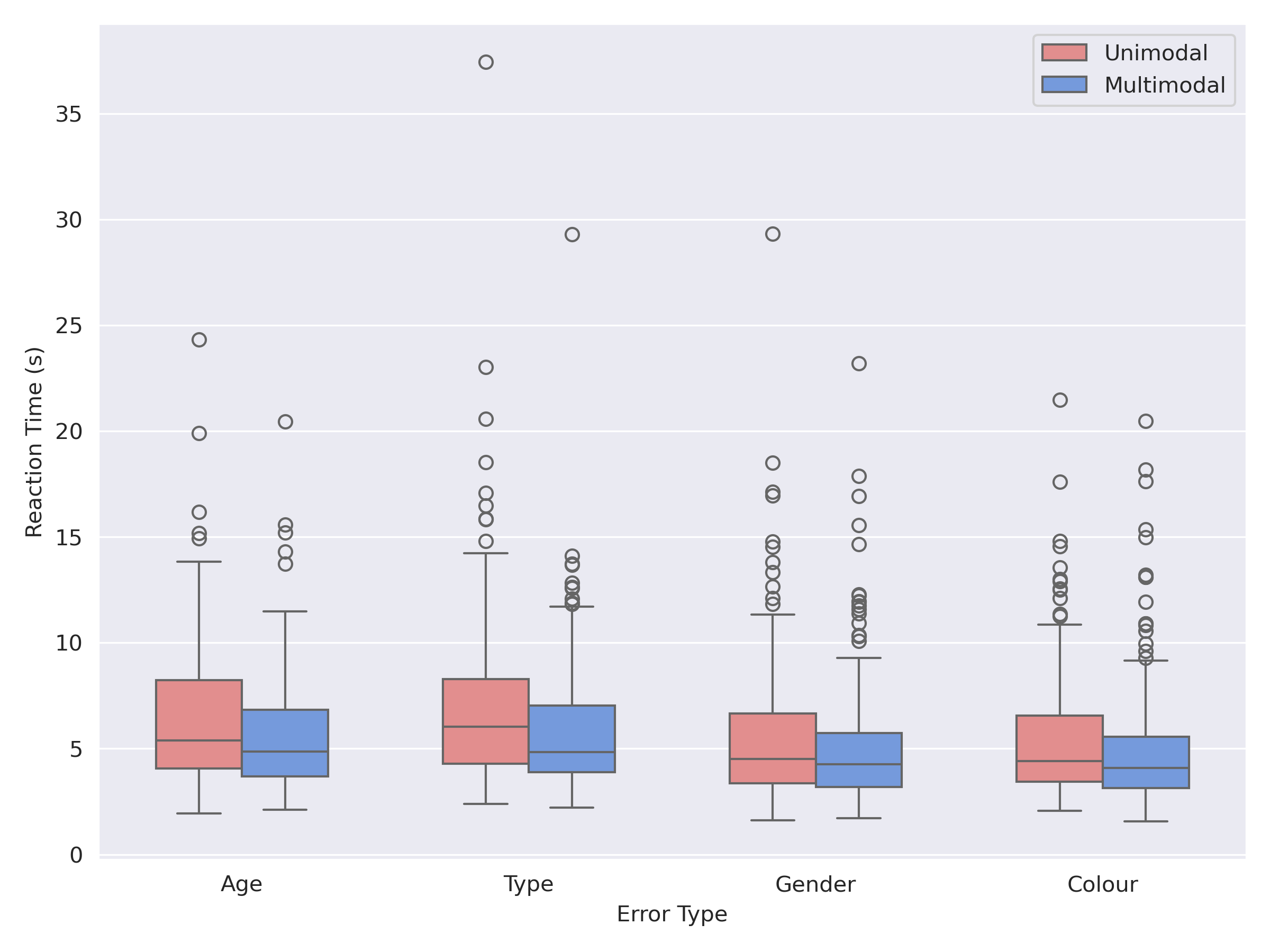}
    \caption{The box plot of the distribution of reaction times for participants across each error type and condition.}
    \label{fig:reaction}
\end{figure}

Figure~\ref{fig:reaction} shows the distribution of reaction times for participants across each error type and condition. An LMM analysis indicates that both error type and modality have a significant effect on reaction time ($p<.001$). The presence of visual context reduces participants' reaction times, suggesting that multimodal inputs enable participants to judge the quality of the texts more quickly. We also observe that participants judge gender and colour errors, which they consider more severe than the other errors, more quickly.

The interaction effect between these two factors is insignificant ($p=.601$), indicating that although humans judge gender and colour errors faster, this is not due to the presence of images.

\subsection{LLM Performance}

Hypothesis $\mathcal{H}_3$ predicted that all multimodal and unimodal LLMs are able to replicate the severity ranking of error types. However, the results in Figure~\ref{fig:score} show that only Doubao in the unimodal condition fully replicates the order in Equation~\ref{eq:order_2}.

GPT-4o, DeepSeek-V3 and DeepSeek-R1 have a slightly different ranking: 
$$\text{\textsc{gender}} \prec \text{\textsc{age}}  \prec \text{\textsc{colour}} \prec \text{\textsc{type}}.$$
Unlike humans, these three LLMs assign low scores to gender errors but high scores to colour errors. Humans perceive colour and gender errors as the most severe for different reasons. The severity of colour errors is due to the fact that colour features are processed by distinct neural mechanisms in the human brain, compared to other object properties~\citep{connell2007representing,liu2024mental}. In contrast, the severity of gender errors is largely influenced by social norms~\citep{ashrafova2024language}. The results suggest that these three LLMs are likely to have learned social norms but have not captured the specialised neural processing mechanisms associated with colour perception in humans.

Doubao exhibits the same error severity ranking as humans in the unimodal condition. In the multimodal condition, it assigns slightly lower quality scores to type errors compared to age errors, but this difference is not statistically significant (see further details in later statistical analyses). Given that this difference is also insignificant in the human experiment, we can conclude that among the 4 LLMs we tested, only Doubao replicates the human error severity ranking.

\paragraph{Effects of Error Type and Modality.}

For the two multimodal LLMs, GPT-4o and Doubao, the results show that they assign significantly higher quality scores to unimodal inputs than to multimodal inputs. We conducted the same statistical analysis as in the human experiment, using a Linear Mixed-Effects Model (LMM). The outcomes in Table~\ref{tab:effect} suggest that, similar to humans, both the error type and the presence of visual context have significant main effects on Doubao, and there is also a significant interaction effect. For GPT-4o, although the main effects of error type and visual context are significant, no interaction effect is observed. This suggests that showing GPT-4o images decreases the scores it assigns, but the change is consistent across different error types.

The estimated fixed effects in Table~\ref{tab:fee} show similar results. Presenting images to the LLMs generally decreases the scores they assign, with almost no interaction effect observed for GPT-4o and very small, statistically insignificant interaction effects for Doubao. Recall that for humans, the presence of visual context significantly reduces their ratings for colour and type errors. Therefore, the expectation of hypothesis $\mathcal{H}_4$ does not hold: the presence of visual context does not influence multimodal LLMs in the same way it affects human judgments.

As for the two unimodal LLMs, DeepSeek-V3 and DeepSeek-R1, we tested only the effect of error type, which, as expected, is significant ($p<.001$).

\paragraph{Pairwise Comparisons.}

We conducted pairwise comparisons between error types, as in the human experiment, and report the results in Table~\ref{tab:post_hoc}. For GPT-4o, as mentioned earlier, it assigns higher quality scores to colour errors compared to humans. This leads to differences between age and colour errors that are in the opposite direction to those observed in humans in both unimodal and multimodal conditions, although these differences are not statistically significant. All other differences align with the direction observed in human judgments and are statistically significant.

For Doubao, since its severity ranking largely aligns with humans, the only inconsistency in terms of the direction of difference is between age and type errors. Nonetheless, since both Doubao and humans consider this difference to be statistically insignificant, this mismatch can be considered acceptable. Another issue with Doubao is that, although it ranks error severity in the same order as humans, it perceives most of the differences between errors as statistically insignificant. As a result, while Doubao may assign scores that follow a similar trend to human judgments, using Doubao as a judge for NLG evaluation could lead to different conclusions, as it regards many types of errors as statistically indistinguishable.

The unimodal LLMs, DeepSeek-V3 and DeepSeek-R1, exhibit a similar pattern to GPT-4o: the difference between age and colour errors is insignificant and in the opposite direction to that of humans. As aforesaid, this is likely because these models have not learned the specialised processing mechanisms for colour that are present in the human brain, leading them to assign overly high quality scores to descriptions with colour errors. All other differences align with the direction observed in human judgments and are statistically significant.

\subsection{Correlation Analysis}

\input{table/correlation}

Analogue to many studies about LLM-as-a-judge, we were also curious about how well the quality assigned by LLMs aligns with that of humans in our controlled experiments. We, thus, computed the Pearson and Spearman correlation coefficients not only between scores by LLMs and humans in the same condition, but also between scores by humans in the multimodal condition and LLMs in the unimodal condition to investigate whether one can use unimodal LLMs (which is way cheaper than multimodal LLMs) to evaluate multimodal NLG. Table~\ref{tab:correlation} charts the correlation coefficients. \footnote{The correlations between the quality scores assigned by LLMs in the multimodal condition and those assigned by humans in the unimodal condition are crossed off, as using an LLM in the multimodal condition to replicate human rating behaviours for unimodal NLG is practically irrelevant.}

For the two multimodal LLMs, Doubao aligns more closely with humans than GPT-4o in both unimodal and multimodal conditions. As expected, each LLM aligns better with humans in the unimodal condition when evaluating unimodal inputs compared to evaluating multimodal inputs. However, it is surprising that the LLM in the unimodal condition aligns better with humans in the multimodal condition than the LLM in the multimodal condition itself. This may suggest that, when evaluating multimodal NLG using the ``LLM-as-a-judge'' paradigm, it might be more effective to provide the LLM with only the textual inputs.

What is even more surprising is that the unimodal LLMs, DeepSeek-V3 and DeepSeek-R1, which have never been trained with multimodal inputs, align better with humans than the multimodal LLMs in both unimodal and multimodal conditions. The results also indicate that DeepSeek-V3 outperforms DeepSeek-R1, suggesting that reasoning ability does not necessarily enhance an LLM's capacity to evaluate NLG outputs.



%% file: table/post_hoc.tex
\begin{table*}[t]
\resizebox{\textwidth}{!}{%
\small
\centering
\begin{tabular}{llccccccccccc}
\toprule
 & & \multicolumn{3}{c}{Human}    & \multicolumn{3}{c}{GPT-4o}   & \multicolumn{3}{c}{Doubao} & DeepSeek V3 & DeepSeek R1\\
\cmidrule(lr){3-5} \cmidrule(lr){6-8}\cmidrule(lr){9-11}\cmidrule(lr){12-12}\cmidrule(lr){13-13}
Type 1 & Type 2 & Both & Uni. & Mul. & Both & Uni. & Mul. & Both & Uni. & Mul. & Uni. & Uni. \\
\midrule
Age & Colour & \makecell[c]{16.27\\****} & \makecell[c]{12.45\\****} & \makecell[c]{20.09\\****} & \colorbox{red}{\makecell[c]{-1.67\\$\times$}} & \colorbox{red}{\makecell[c]{-1.25\\$\times$}} & \colorbox{red}{\makecell[c]{-2.08\\$\times$}} & \makecell[c]{7.29\\**} & \colorbox{yellow}{\makecell[c]{4.17\\$\times$}} & \makecell[c]{10.42\\**} & \colorbox{red}{\makecell[c]{-1.04\\$\times$}} & \colorbox{red}{\makecell[c]{-4.38\\$\times$}} \\
Age & Type & \makecell[c]{-4.53\\**} & \makecell[c]{-8.43\\****} & \makecell[c]{-0.63\\$\times$} & \makecell[c]{-10.94\\****} & \makecell[c]{-8.96\\**} & \makecell[c]{-12.92\\****} & \colorbox{red}{\makecell[c]{0.63\\$\times$}} & \colorbox{yellow}{\makecell[c]{-3.33\\$\times$}} & \colorbox{green}{\makecell[c]{4.58\\$\times$}} & \makecell[c]{-8.33\\****} & \makecell[c]{-10.83\\****} \\
Age & Gender & \makecell[c]{26.41\\****} & \makecell[c]{25.25\\****} & \makecell[c]{27.57\\****} & \makecell[c]{8.54\\***} & \makecell[c]{7.92\\**} & \makecell[c]{9.17\\***} & \makecell[c]{17.92\\****} & \makecell[c]{22.08\\****} & \makecell[c]{13.75\\****} & \makecell[c]{15.63\\****} & \makecell[c]{13.33\\****} \\
Colour & Type & \makecell[c]{-20.79\\****} & \makecell[c]{-20.88\\****} & \makecell[c]{-20.71\\****} & \makecell[c]{-9.27\\****} & \makecell[c]{-7.71\\**} & \makecell[c]{-10.83\\***} & \makecell[c]{-6.67\\**} & \colorbox{yellow}{\makecell[c]{-7.50\\$\times$}} & \colorbox{yellow}{\makecell[c]{-5.83\\$\times$}} & \makecell[c]{-7.29\\***} & \makecell[c]{-6.46\\**} \\
Colour & Gender & \makecell[c]{10.14\\****} & \makecell[c]{12.8\\****} & \makecell[c]{7.49\\****} & \makecell[c]{10.21\\****} & \makecell[c]{9.17\\**} & \makecell[c]{11.25\\***} & \makecell[c]{10.63\\****} & \makecell[c]{17.92\\****} & \colorbox{yellow}{\makecell[c]{3.33\\$\times$}} & \makecell[c]{16.67\\****} & \makecell[c]{17.71\\****} \\
Type & Gender & \makecell[c]{30.93\\****} & \makecell[c]{33.67\\****} & \makecell[c]{28.20\\****} & \makecell[c]{19.48\\****} & \makecell[c]{16.88\\****} & \makecell[c]{22.08\\****} & \makecell[c]{17.29\\****} & \makecell[c]{25.42\\****} & \makecell[c]{9.17\\**} & \makecell[c]{23.96\\****} & \makecell[c]{24.17\\****} \\
\bottomrule
\end{tabular}}
\caption{Estimated differences for the pairwise comparisons between different error types. `Uni.' means the unimodal condition, and `Mul' means the multimodal condition. Significance Marking: $^{***} p<.001$; $^{**} p<.01$; $^{*} p<.05$; $\times p\leq.05$. The results highlighted in red represent differences that are in the opposite direction to those of humans. The results highlighted in yellow represent differences that align with the direction of human judgments but are not statistically significant. The results highlighted in green represent differences that are in the opposite direction to those of humans, but all of these differences are statistically insignificant.}
\label{tab:post_hoc}
\end{table*}

%% file: table/correlation.tex
\begin{table}[t]
\resizebox{\columnwidth}{!}{%
\centering
\begin{tabular}{lcccc}
\toprule
& \multicolumn{2}{c}{Human Uni.}    & \multicolumn{2}{c}{Human Mul.}\\
\cmidrule(lr){2-3} \cmidrule(lr){4-5}
& Pearson & Spearman & Pearson & Spearman \\
\midrule
GPT-4o Uni. & .7232 & .7789 & .5951 & .6674 \\
GPT-4o Mul. & \sout{.4315} & \sout{.4259} & .5992 & .5062 \\
Doubao Uni. & .7391 & \textbf{.8305} & .6662 & .7451 \\
Doubao Mul. & \sout{.3652} & \sout{.2357} & .6801 & .5439 \\
DeepSeek V3 & \textbf{.7762} & .8246 & .6700 & \textbf{.7663} \\
DeepSeek R1 & .7561 & .7492 & \textbf{.6902} & .7529 \\
\bottomrule
\end{tabular}}
\caption{The correlations between the quality scores assigned by humans and those assigned by LLMs in both unimodal and multimodal conditions are presented.}
\label{tab:correlation}
\end{table}


%% file: section/discussion.tex
\section{Discussion: Which LLM is the ``Best''?}

So far, we have replicated the human experiments on multiple LLMs and analysed these models from various perspectives. The final question is which LLM is best at handling error severity in NLG. The answer to this question is two-fold: first, which LLM most accurately replicates human behaviours, and second, which LLM is best suited to serve as an evaluator for NLG tasks, particularly for image captioning.

Regarding the first question, the winner is Doubao. It is the only LLM that fully replicates the ranking of error severity observed in humans. All other models assign high scores to colour errors, likely because they have not learned the specialised processing mechanisms for colour features.

However, using Doubao as an NLG evaluator may lead to very different conclusions compared to human evaluation. This is because Doubao tends to judge different types of errors as equally severe. The best choice for an NLG evaluator is DeepSeek-V3. As a unimodal model, it achieved the highest correlation with human judgments in both the unimodal and multimodal conditions. It is the most cost-effective LLM among those we examined. The only drawback is that it may struggle when many colour errors are present in NLG outputs.

%% file: section/appendix.tex
\section{Examples of Each Error Type}
\label{sec:appendix_example}

For the image in Figure~\ref{fig:example}, we have the following reference description and erroneous descriptions:
\begin{itemize}
    \item 一位男人穿着黄色上衣在网球场打网球。 \\
    `A man in a yellow shirt plays tennis on the tennis court.'
    \item Gender Error: 一位\textcolor{red}{女}人穿着黄色上衣在网球场打网球。
    \item Age Error: 一位男\textcolor{red}{孩}穿着黄色上衣在网球场打网球。
    \item Type Error: 一位男人穿着黄色\textcolor{red}{大}衣在网球场打网球。
    \item Colour Error: 一位男人穿着\textcolor{red}{紫}色上衣在网球场打网球。
\end{itemize}

\section{Criteria for Image Selection} \label{sec:appendix_criteria}

\begin{enumerate}
    \item They should be full-colour images.
    \item There should be a human protagonist, with
their face and at least half their body visible.
    \item The content of the images should be clearly
recognisable.
    \item Each clothing item should have a single colour.
    \item Clothing items should have different colours.
\end{enumerate}

\section{Prompts for the LLM Experiment} \label{sec:appendix_prompt}

For unimodal stimuli, we used the prompt in Table~\ref{tab:prompt_uni} to guide both unimodal and multimodal LLMs in rating these stimuli. For multimodal stimuli, we used the prompt in Table~\ref{tab:prompt_mul} to instruct multimodal LLMs to rate the stimuli.

\begin{table}[htbp]
    \centering
    \small
    \begin{tabular}{p{0.95\linewidth}}
        \toprule
你是一位图像描述系统的质量评估专家。\\
\\
现在你将看到两段文字：\\
第一段是某张图像的正确描述（作为参考）\\
第二段是图像描述系统生成的待评估描述\\
请你根据参考描述的内容，对系统生成的描述在整体质量上的表现进行主观评分（1-100 分）。\\
打分时请考虑描述与参考描述之间在语义上的接近程度、信息覆盖、表达合理性等因素，但不需细分维度。\\
\\
请严格按照以下格式作答：\\
分数：[0-100] \\
理由：[打分的理由] \\
\\
输入：\\
正确描述：\{Reference Description\}\\
待评估描述：\{Description\}\\
        \bottomrule
    \end{tabular}
    \caption{\(Prompt\) used for Instructing LLMs to rate Unimodal Stimuli.}
    \label{tab:prompt_uni} 
\end{table}

\begin{table}[htbp]
    \centering
    \small
    \begin{tabular}{p{0.95\linewidth}}
        \toprule
            你是一位图像描述系统的质量评估专家。\\
            \\
现在你将看到一张图片和系统为这张图片生成的一段文字描述。\\
请你从专业角度出发，依据整体质量对这段描述进行0到 100 分的打分。\\
\\
评分标准仅基于你对描述整体质量的主观判断，而不需要从多个维度进行细分分析。\\
\\
请严格按照以下格式作答：\\
分数：[0-100]\\
理由：[打分的理由]\\
\\
描述：\{Description\}\\
\{Image\}\\
        \bottomrule
    \end{tabular}
    \caption{\(Prompt\) used for Instructing LLMs to rate Multimodal Stimuli.}
    \label{tab:prompt_mul} 
\end{table}
